\documentclass[10pt, a4paper]{article}
\usepackage{lrec2000}
\usepackage{epsfig,url,boxedminipage,alltt}
\def\sburl#1{[{\small\url{#1}}]}
\def\smtt#1{{\small\tt #1}}
\def\boxfig#1{\fbox{\parbox{0.97\linewidth}{\centerline{\epsfig{#1}}}}}
\newenvironment{sv}{\scriptsize\begin{alltt}}{\end{alltt}\normalsize}
\newtheorem{defn}{Definition}

\title{ATLAS: A Flexible and Extensible Architecture for Linguistic
Annotation}

\name{
  Steven Bird$^{\ast}$,
  David Day$^{\dagger}$,
  John Garofolo$^{\ddagger}$,\\
  John Henderson$^{\dagger}$,
  Christophe Laprun$^{\ddagger}$ and
  Mark Liberman$^{\ast}$
}

\address{
$^{\ast}$Linguistic Data Consortium, University of Pennsylvania,
  3615 Market Street, Philadelphia, PA 19104, USA\\
$^{\dagger}$MITRE Corporation, 202 Burlington Road,
  Bedford, MA 01730, USA\\
$^{\ddagger}$National Institute of Standards and Technology,
  100 Bureau Drive, Mailstop 8940, Gaithersburg, MD 20899-8940, USA
}

\abstract{
We describe a formal model for annotating linguistic artifacts, from
which we derive an application programming interface (API) to a suite
of tools for manipulating these annotations.  The abstract logical
model provides for a range of storage formats
and promotes the reuse of tools that interact through this
API.  We focus first on ``Annotation Graphs,'' a graph model for
annotations on linear signals (such as text and speech)
indexed by intervals, for which efficient database storage and
querying techniques are applicable.  We note how a wide range of
existing annotated corpora can be mapped to this annotation graph
model.  This model is then generalized to encompass a wider variety of
linguistic ``signals,'' including both naturally occuring phenomena (as
recorded in images, video, multi-modal interactions, etc.), as well as
the derived resources that are increasingly important to the
engineering of natural language processing systems (such as word
lists, dictionaries, aligned bilingual corpora, etc.).  We conclude
with a review of the current efforts towards implementing key pieces
of this architecture.
}

\begin{document}

\maketitleabstract

\section{Introduction}

Annotated corpora are a central component of research in human
language technology.
As corpora have proliferated across a rapidly
expanding set of languages, disciplines and technologies, the lack of
agreed standards has become a critical problem.  The standardization
of tagsets is necessarily an open-ended task, and will always be subject to
revision as the underlying domains change and the theories evolve.
Yet with no agreed data models and application programming interfaces
(APIs), the bazaar of tools and formats continues to expand, and
incompatibilities proliferate.  Adapting existing annotation tools to
new formats often requires non-trivial re-engineering.  Seen from this
perspective, general-purpose annotation tools and formats are a
distant prospect.

In recent work, Bird and Liberman \shortcite{BirdLiberman99}
have demonstrated commonality across
a diverse range of annotation practice.  Existing tools generally
implement a two-level architecture consisting of an application level
(the interface to a user or to external software)
and a physical level (the storage format).  It is
possible to interpose an intermediate, logical level, which is
independent of the application and the physical storage (cf. the three-level
architecture for relational databases).  Once this step is taken,
wide-ranging integration of tools and formats becomes possible.  The
logical level we will propose in this paper
is based around the notion of an ``annotation
graph,'' which is a labeled, directed acyclic graph with time-stamps on
some of its nodes, anchoring the graph to a physical signal.
Annotation becomes the fundamental act of associating a symbolic
property (the arc's label) to an extent of signal data (the arc's time span).

ATLAS: ``Architecture and Tools for Linguistic Analysis Systems'' is a
recent initiative involving NIST, LDC and MITRE, arising from an
array of applications needs spanning corpus construction,
evaluation infrastructure, and multi-modal visualization.
The principal goal of ATLAS is to provide powerful abstractions over
annotation tools and formats in order to maximize flexibility and
extensibility.  Our approach has been to isolate and abstract over
the physical and logical levels of annotation tools and formats,
leaving application- and domain-specific issues to the side.
The abstract physical level is a persistent XML representation for
long-term storage, exchange, and pipelining, and this level is
called ATLAS Interchange Format (AIF).
The abstract logical representation is the internal representation
for broad classes of data.  It includes linear signals (text, speech)
indexed by intervals (i.e. annotation graphs),
images indexed by bounding boxes, and
additional generic representations for other data classes
(lexicons, tables, aligned corpora).

This paper is structured into three main parts.  We begin by
describing the problem and our approach, then describe the open
architecture.  Next, we present the annotation graph model and our
generalization of it to higher dimensions, before reporting
progress on the implementation.  We conclude with a discussion of
future plans and an invitation for wider participation.

\section{The Bazaar of Tools and Formats}

\begin{figure}
\boxfig{figure=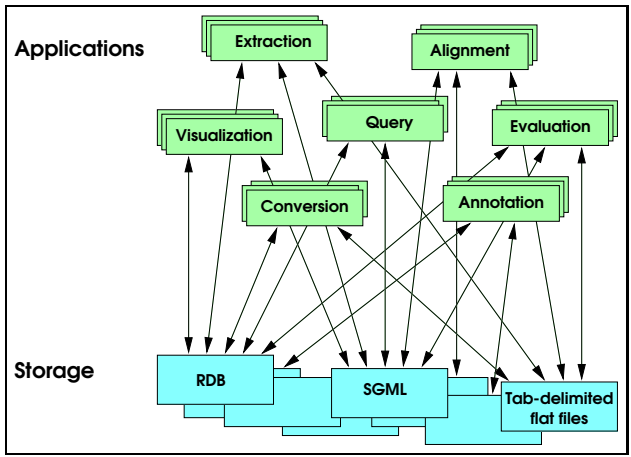,width=\linewidth}
\caption{The Two-Level Architecture}
\label{fig:2level}
\end{figure}

Over the last several years, there has been a veritable bloom in new
language technology research projects. Measurable progress is being
made in many core language technologies including automated speech
recognition, information extraction, information retrieval, machine
translation, natural language processing and others.  To support the
development and evaluation of these technologies, there has also been
a boom in the development of language research resources.

Not only are more of these projects being formed, they typically
adopt a shorter development cycle than has been customary in the past.
While much of this acceleration is due to the availability of
greatly increased processor speeds and storage capacities, these
projects are also benefiting from the creation of spin-off projects
focusing on integrated multi-component applications. The result is
that many researchers are beginning to develop toolkit approaches to
language technology development.

Unfortunately, to date, these efforts have almost always been
intra-domain. Likewise, the research corpora that have been developed
for these projects have been created and deployed with minimal
consideration toward re-use and extensibility. We therefore live in a
world with an increasing variety of technology capabilities and a
corresponding wealth of research corpora, yet these capabilities
and corpora are domain-specific, task-specific, and often even
site-specific (see Figure~\ref{fig:2level}).

\begin{table}[ht]
\begin{center}
\begin{tabular}{|c|l|l|}
\hline
\bf Task & \bf Formats    & \bf Tools \\ \hline\hline
Hub-4    & UTF (SGML)     & Hub-4 Transcription tools\\
ASR      & STM, CTM       & SCLITE \\ \hline
IE-ER    & UTF+           & Alembic Workbench\\
         &                & NE99NS\\ \hline
TDT      & Raw text, RDB  & TDT Event annotation tools\\
         & Eval indexes   & TDTEval \\
         & TDT lists      & \\ \hline
SDR      & raw text, SRT  & TREC tools \\
         & SGML lists     & TREC\_EVAL \\
         & TREC qrels     & \\ \hline
ACE      & Raw text + ASR & Alembic Workbench\\
         & + OCR text     & EDT\_REF\_COMPARE\\
         & APF            & \\
         & ATLAS AIF?     & \\ \hline
\end{tabular}
\caption{Broadcast News Tasks, Formats and Tools}
\label{table:bn}
\end{center}
\end{table}

For example, for several years, NIST has been organizing and
implementing language technology evaluations for several research
domains using recordings and transcriptions of radio and television
news broadcasts (Table~\ref{table:bn}). Although all of these
evaluations have used basically the same source data, each required
the development of customized formats and tools. Two years ago, NIST
made an attempt to unify its transcription formats with the SGML-based
Universal Transcription (UTF) format \cite{UTF98} for the DARPA Hub-4
ASR evaluations \cite{Pallett99}. But, in 1999, when the Hub-4
evaluation was expanded to include an entity recognition evaluation
\cite{Przybocki99}, NIST found that the UTF format could not
accommodate the different tokenization schemes required by the ASR and
extraction communities. It was at this point that NIST realized the
need for a more abstract, open-ended transcription format that could
accommodate such unforeseen changes. Such a format would have to been
domain-independent and permit any conceivable extension.

NIST believed that the best solution to the formats dilemma could be
achieved via a multi-site effort initially including NIST, the LDC,
and MITRE.  NIST would contribute its expertise in language technology
evaluation and infrastructure; the LDC would contribute its expertise
in corpus development including its recent research in abstract
annotation representations using annotation graphs,
and its involvement with the new ISLE and TalkBank projects
\cite{ISLE,TalkBank}, and MITRE would contribute its expertise in language
technology development and annotation and visualization using the
Alembic Workbench \cite{Day97}.  Thus, the ATLAS working group was formed.

The group's mission was to develop a general
architecture for annotation including a logical data format, an API
and tool-set, and a persistent data representation.  The architecture
would, by definition, need to be modular, flexible and extensible.
First, this architecture would facilitate the exchange and reuse of
existing language resources.  Such resources would be moved in and out
of the ATLAS framework via conversion routines.  As such, ATLAS
could also act as an interlingua for language corpora.
Further, the architecture would provide a platform for the
extension of such corpora and the development of new ATLAS-native
resources.  The ATLAS philosophy would eschew the tradition of
imposing monolithic applications or formats on the language research
community.  Rather, ATLAS would serve as a conduit to enable the greater
flow of language resources throughout the language research community.

\section{The ATLAS Architecture}

\begin{figure}[b]
\boxfig{figure=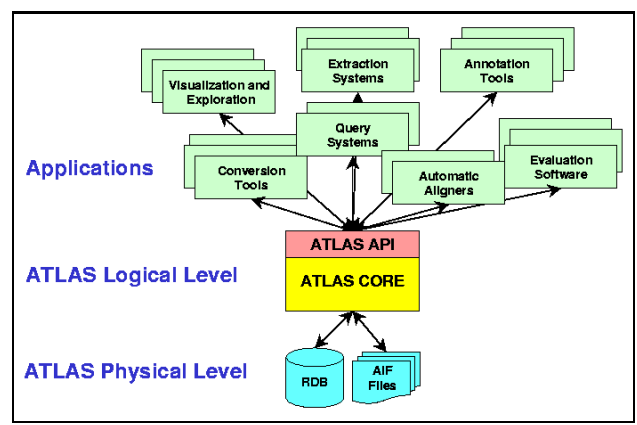,width=\linewidth}
\caption{ATLAS Layered Solution}
\label{fig:3level}
\end{figure}

ATLAS consists of three levels: application, logical, and physical.
The overall structure is depicted in Figure~\ref{fig:3level}.
We discuss each of these layers in turn, beginning with the
middle, logical level.

\subsection{The logical level}

The logical layer consists of a linguistic formalism and an API.  The
formalism is the annotation graph model and its generalization to
higher-dimensional cases.  The formalism is a generalization of
the annotation graph model, called annotation sets.
The API defines a set of procedures for creating, modifying,
searching and storing well-formed annotation sets.

\subsection{The physical level}

The API specification will allow for a multiplicity of physical
storage implementations that applications are free to access in
various ways -- via networked client-server models, or via
libraries linked directly into application binaries,
or via scripting languages.  The two dominant
storage strategies that we are implementing are:
ATLAS Interchange Format (AIF), an XML interchange format; and
an RDBMS accessible from ODBC-compliant calls.

The AIF XML format will provide a simple, wide-coverage interchange
format for which conversion programs to and from other annotation
formats are being developed.  The AIF XML annotations are ``stand-off''
in the sense that they reference
the signal being annotated (whether text or speech, or
some other modality).  This considerably simplifies the encoding of
multiple layers of annotation, especially those that would involve
crossing brackets if they were embedded in the signal data.
Given the generality of the annotation graph formalism, we
believe that AIF could serve as an interlingua for sharing annotated
linguistic corpora among language processing applications, reducing
the number of conversion tools that need to be written.

The RDBMS implementation will provide efficient access to large,
heterogeneous linguistic databases, and open the door to analysis
of a nature and scale that was hitherto impractical or impossible.

\subsection{The application level}

This level contains a rich diversity of components.
The annotation set formalism, as implemented in ATLAS,
will reduce the burden on language engineering
applications development.  To demonstrate this claim, we are
developing a range of initial applications, as described below.
The modularity provided by the API works for
applications as well: distinct components will communicate
their operations on annotations via the API, greatly
enhancing the re-usability of application-level components.

\section{The Annotation Graph Model}\label{sec:ag}

Annotation graphs were presented by Bird and Liberman
\shortcite{BirdLiberman99} as follows.

\begin{defn}
An \textbf{annotation graph} $G$ over a label set $L$ and
timelines $\left<T_i, \leq\right>$ is a 3-tuple
$\left< N, A, \tau \right>$ consisting of a node set $N$,
a collection of arcs $A$ labeled with elements of $L$,
and a time function $\tau$, which satisfies the following conditions:

\begin{enumerate}\setlength{\itemsep}{0pt}

\item $\left< N, A \right>$ is a labeled acyclic digraph
  containing no nodes of degree zero;

\item $\tau: N \rightharpoonup \bigcup T_i$,
  such that, for any path from node $n_1$ to $n_2$ in $A$,
  if $\tau(n_1)$ and $\tau(n_2)$ are defined, then
  $\tau(n_1) \leq \tau(n_2)$;

\end{enumerate}
\end{defn}

Note that annotation graphs may be disconnected or empty, and that they must
not have orphan nodes.

\begin{figure}
\begin{boxedminipage}[t]{\linewidth}
\begin{sv}
train/dr1/fjsp0/sa1.wrd:   train/dr1/fjsp0/sa1.phn:
2360 5200 she              0 2360 h#
5200 9680 had              2360 3720 sh
9680 11077 your            3720 5200 iy
11077 16626 dark           5200 6160 hv
16626 22179 suit           6160 8720 ae
22179 24400 in             8720 9680 dcl
24400 30161 greasy         9680 10173 y
30161 36150 wash           10173 11077 axr
36720 41839 water          11077 12019 dcl
41839 44680 all            12019 12257 d
44680 49066 year           ...
\end{sv}
\centerline{\epsfig{figure=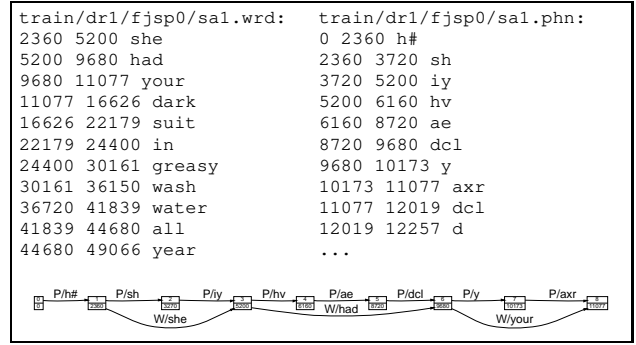,width=0.95\linewidth}}
\end{boxedminipage}
\caption{TIMIT Annotation Data and Graph Structure}\label{fig:timit}
\end{figure}

The formalism can be illustrated with an application to a simple
speech database, the TIMIT corpus of read speech \cite{TIMIT86}.
This was the first speech database to be widely distributed.  It contains
broadband recordings of 630 speakers of 8 major dialects of American
English, each reading 10 phonetically rich sentences
\sburl{www.ldc.upenn.edu/Catalog/LDC93S1.html}.
Figure~\ref{fig:timit} shows part of the annotation of one of the
sentences.  The file on the left contains word transcription, and
the file on the right contains phonetic transcription.  Part of
the corresponding annotation graph is shown underneath.

In Figure~\ref{fig:timit}, each node displays the node identifier
and the time offset (in 16kHz sample numbers).  The arcs are decorated
with type and label information.  The type \smtt{W} is for words and
the type \smtt{P} is for phonetic transcriptions.

\begin{figure}[b]
\boxfig{figure=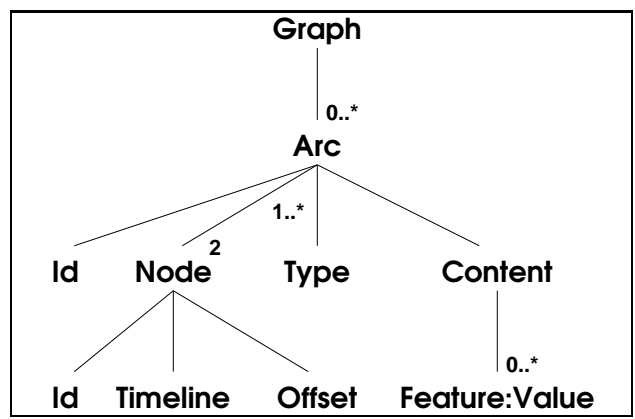,width=0.9\linewidth}
\caption{The Annotation Graph Object Model}
\label{fig:model}
\end{figure}

A simplified chart showing the primary objects and their relationships
is shown in Figure~\ref{fig:model}.  A graph object is a collection
of zero or more arc objects, where these specify an identifier, two
nodes, a type, and the content.  A node specifies an identifier,
a timeline, and an offset into that timeline.  A timeline simply
denotes a set of signals that share the same abstract notion of time
\cite{BirdLiberman99}.
For example, multichannel audio signals, or aligned video and audio data
are signal sets which have a single time-base, and which would be
grouped together by a common timeline.  The content object consists
of a set of feature-value pairs.  The object identifiers are intended
to support incoming references from external resources, and cross-references
within a particular annotation graph.  For example, the value of some
feature of a arc could be the identifier of another arc.

The API for this model includes such functions as
\smtt{Arc::insertArc(n1,n2,t,l)} to insert an arc from
node \smtt{n1} to node \smtt{n2} with type \smtt{t}
and label \smtt{l}, and
\smtt{Arc::splitArc(a)} to replace a single arc with
a path of two arcs spanning the same nodes.

\section{Generalizing the Model}
\label{sec:gen}

Annotation graphs are useful for annotating linear data types,
including intervals found in any particular dimension of a variety of
sequentially-arranged signals.  However, there are more signal types
which we can envision annotating, using the same basic philosophy of
indication and characterization.  Below we give a generalization of
annotation graphs to encompass a class of signal types which
correspond to $n$-dimensional vector spaces.

\subsection{Higher-dimensional cases}

There are many signals having dimension greater than one,
that we acquire in our computers, and that
are amenable to linguistic modeling techniques.
The input signal to optical character recognition (OCR) is
one example of a signal that loses (or at least obfuscates) its
temporal dimension prior to acquisition by computer.  When annotating
the communicative content in a sign language video, one may want to
highlight the portion of each of a sequence of images that is involved
in the signing, and the same should be done in a lip-reading
experiment where the input stream is television broadcast or movie,
for example.  ATLAS strives to model the annotations required in both
reference and hypothesis data for an OCR system, a sign language
recognition system, lip-reading system, as well as all well-formed
combinations of these.

Annotation graphs, however, do not provide a natural framework for
identifying regions of a signal having more than one dimension.
Intuitively, we want to be able to specify bounding boxes, or other
kinds of bounding areas, as a target for further characterization in
the two dimensions of OCR $(x,y)$.  The analogy carries for higher
dimensions such as video $(x,y,t)$ where we want bounding volumes.
The generalization of annotation graphs that facilitates this is
Cartesian Annotation Sets.  The name is chosen thus because
we want to select a region of a vector space, rather than just
an interval.  For brevity we will usually refer to these simply
as ``annotation sets.''  Full details are given below.

There are further generalizations of annotation being explored, as one
can imagine targeting signals which inhabit spaces that are not
results of repeated Cartesian products (e.g. trees, partially ordered
sets, relational or semistructured data).
The annotation sets cover a very large set of signals that are captured
by current digitizing equipment, however, and they are a primary
target for development in ATLAS.

\subsection{Annotation sets}

In the terminology of annotation graphs, a labeled arc
relates symbolic label data to an extent of signal, and this extent is
specified using a pair of nodes.  The arc serves a dual purpose: as an
ordered pair of nodes specifying an extent; and as an entity which can
accommodate label data.  We separate these functions by replacing nodes
and labeled arcs with some new constructs which reduce to nodes and
labeled arcs in the linear case.

We begin by adopting ``region'' as a term for any extent of signal,
regardless of dimensionality.  In the one-dimensional case, regions
are simply intervals, and there are various ways to specify these
(endpoints, start-point plus offset, midpoint plus radius).  The
endpoint method is convenient since the boundaries are typically
shared by several regions, and changing an endpoint value does not
then require any further propagation of information.
In the two-dimensional case we may specify regions using points or line
segments; we select ``anchor'' as the cover term.
Again, the choice of the anchoring method
is based on the need to share boundaries between regions.
The terminology extends to the three dimensional case,
and is neutral as to whether the dimensions are spatial or temporal.
We define an ATLAS ``annotation'' to be a relation
between such regions and (structured) labels.

In the one-dimensional case, a collection of arcs is formed into
a graph, representing a set of annotations on a particular signal.
In the generalized model, we refer to a set of annotations as
an ``annotation set''.
A corpus is then a set of annotation sets, and a collection of
corpora (such as LDC-Online) is a set of sets of annotation sets.
If the provenance of any particular annotation is stored in its label,
and if the signal information resides in the region structure, then
there is no harm in flattening this nested set structure.  Therefore
we view ``annotation set'' as a top-level construct.

\section{ATLAS Design and Implementations}

\begin{figure*}[t]
\centering{\epsfig{figure=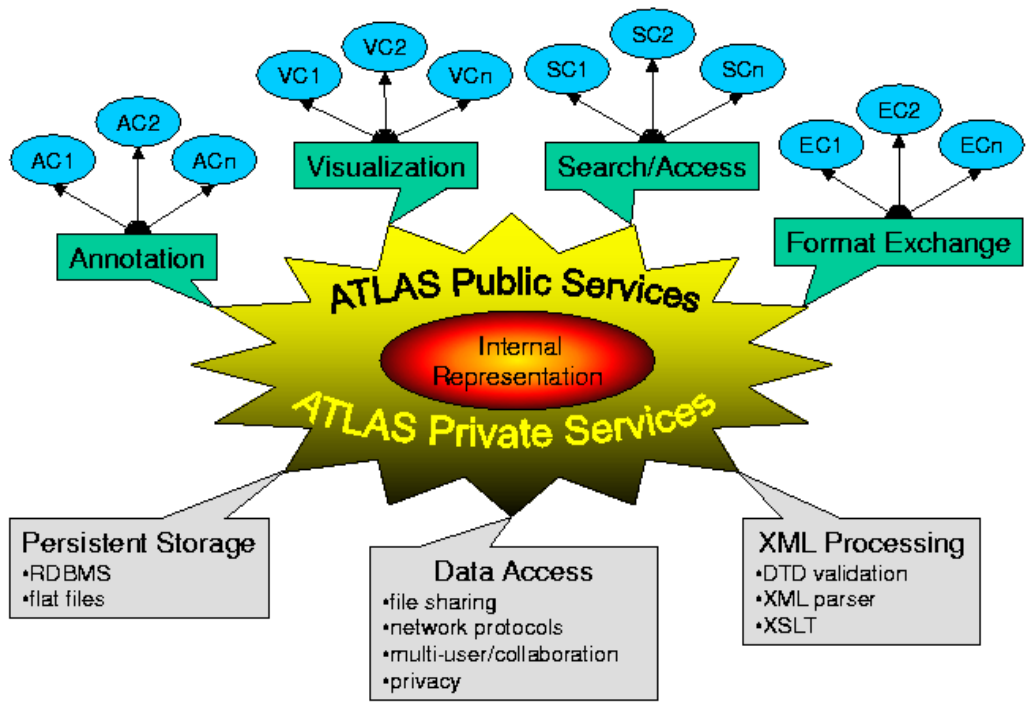,width=.8\linewidth}}
\vspace*{1ex}\hrule
\caption{ATLAS Architecture}
\label{fig:architecture}
\end{figure*}

ATLAS is an open architecture that provides a flexible and extensible
framework for linguistic applications (Figure~\ref{fig:architecture}).
Significant efforts have been put in the design to try to ensure that
the architecture will evolve gracefully as new domains and tasks
arise.  The general philosophy behind ATLAS design consists in trying
to provide a component framework on which new applications can be
built and in which new components can be integrated with ease.

We base our design approach on the object paradigm. With respect to
this approach, we have tried to clearly identify the concepts that
are needed to solve the annotation problem, building on Bird and
Liberman's survey (see \sburl{www.ldc.upenn.edu/annotation/}).
We designed program entities (objects) by
assigning them clear responsibilities and by separating the
interface from the implementation.
This approach has several well-known benefits; here we mention
just three, and describe their significance in the present context.

First, this approach allows us to define APIs to ATLAS
components that are secure and controlled entry points to the whole
architecture.  Second, we can
vary the implementations of components,
offering customization and ``plug'n'play'' capabilities to
developers.  Third,
language-independent interface definitions will permit
ATLAS components to be written using a variety of languages.

\subsection{ATLAS APIs}

The ATLAS APIs are defined to leverage the architecture in an
efficient way. They provide entry points to the functionality offered
by the framework and ensure encapsulation of ATLAS internals,
allowing enhancements to be made to the core functionality without
impacting backwards compatibility.

ATLAS will provide a collection of APIs that address a variety of
issues, built around the core annotation set API.
This core API provides means to create, edit and delete
annotations. It enables users and ATLAS developers to manipulate
annotation components at the lowest level of abstraction.

Another set of components will support the core API and
provide persistent storage capabilities for the architecture. These
components will allow ATLAS-compliant tools to abstract the specifics
of physical storage and transparently interact with flat files or
databases. A complete set of input/output components will be provided
for the ATLAS Interchange Format (see below).

Components will be also developed which provide higher-level services
to ATLAS-compliant tools, easing the task of creating those tools.
These components will also facilitate visualization and
editing of annotations and will support a query interface.
They will allow third-parties to easily and
rapidly develop new ATLAS-compliant tools.

Finally, we are investigating the feasibility of creating components
that will ease inter-component communication, transparent access to
remote databases, and collaborative annotation.

\begin{figure}[ht]
\boxfig{figure=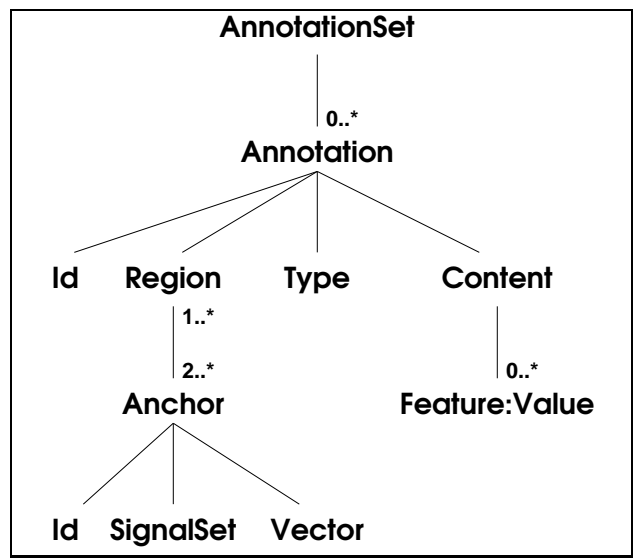,width=0.9\linewidth}
\caption{ATLAS Object Model}
\label{fig:object-model}
\end{figure}

\subsection{The core object model and API}

A simplified chart showing the primary objects
and their relationships is shown in Figure~\ref{fig:object-model}.
As in the annotation graph object model (Figure~\ref{fig:model}),
annotations specify an identifier, a type, and content.
The key difference is that we have abstracted away from
one-dimensional annotation by providing the region object.
The region object references two or more anchors, where these
can define an interval in 1-space or a bounding box or
polygon in 2-space.  An anchor carries an identifier (so that it
may be shared by more than one region), and it specifies a set
of signals (generalizing the notion of timeline).

The API for the ATLAS object model is outlined in Figure~\ref{fig:api},
in simplified form.  For each method the object, method name and
arguments are specified.  Up-to-date information about the API
is available from the ATLAS website.

\begin{figure}[t]
\begin{boxedminipage}[t]{\linewidth}
\begin{list}{}{\setlength{\itemindent}{-\leftmargin}\setlength{\rightmargin}{0pt}\setlength{\itemsep}{0pt}}

\item
  {\sf Anchor::setOffset(Offset)}\\
  Set an anchor's offset to the specified value

\item
  {\sf Anchor::getIncoming()}\\
  Get the incoming annotations to the specified anchor

\item
  {\sf Annotation::setStart(Anchor)}\\
  Set the start anchor of an annotation to the specified anchor

\item
  {\sf Annotation::getStart()}\\
  Access the start anchor of an annotation

\item
  {\sf Annotation::setFeature(feature, value)}\\
  Set the specified feature of the annotation to this value

\item
  {\sf AnnotationSet::add(Annotation)}\\
  Add a new annotation to the collection

\item
  {\sf AnnotationSet::split(Annotation)}\\
  Split an annotation, creating a sequence of adjacent annotations

\item
  {\sf AnnotationSet::remove(Annotation)}\\
  Remove the annotation from the collection

\item
  {\sf AnnotationSet::getAnchorSetByOffset(Offset)}\\
  Get anchors with the specified offset

\item
  {\sf AnnotationSet::getByType(type)}\\
  Get the annotations of type t

\item
  {\sf AnnotationSet::getByFeature(feature, value)}\\
  Get the annotations with feature = value

\item
  {\sf AnnotationSet::getByTimeline(timeline)}\\
  Get annotations of this timeline

\end{list}
\end{boxedminipage}
\caption{A Fragment of the ATLAS API}
\label{fig:api}
\end{figure}

\subsection{ATLAS Interchange Format}

An example of the current instantiation of the ATLAS Interchange
Format is shown in Figure \ref{fig:aif}.  This example shows two
signals that exist outside this annotation file, one a video of someone
called Bill speaking in sign language, and the other an ASCII
file consisting of a transcription of Bill's utterances.  One arc (A1)
identifies a portion of the video signal as containing Bill's signing
of the ASL letter ``e''.  Another arc (A2) identifies a portion of the
text signal as containing a word whose part of speech is ``VBD''.
Finally, a third arc (A3) identifies a larger portion of the video
signal as being transcribed into the word (or, more literally, the
portion of the text signal) identified by the already defined A2 arc.
Notice that the ``content'' of an arc may contain either a set of
attribute/value fields (arc A1), a literal string or symbol (arc A2),
or even a reference to another arc by its unique identifier (arc A3).
The exact syntax of the ATLAS Interchange Format is still being
refined, but this example and the one following it indicate some of
the general components that will be supported.

\begin{figure}
\begin{boxedminipage}[t]{\linewidth}
\begin{sv}
<AnnotationGraph>
<AG_Signal SignalID="S1" 
  Format="video:mpeg-1" ArcTypes="ASL"
  Location="file:bill.signing.mpeg"/>
<AG_Signal SignalID="S2" 
  Format="text:ascii" ArcTypes="NAR"
  Location="file:bill.signing.narrative.cc"/>
<AG_Node NodeId="V0" Signal="S1"
  Offset="382.520" units="Seconds"/>
<AG_Node NodeId="V1" Signal="S1"
  Offset="383.922" units="Seconds"/>
<AG_Node NodeId="V2" Signal="S1"
  Offset="384.731" units="Seconds"/>
<AG_Node NodeId="V3" Signal="S2" 
  Offset="78" units="Characters"/>
<AG_Node NodeId="V4" Signal="S2" 
  Offset="85" units="Characters"/>
<AG_Arc ID="A1" StartNode="V0"
  EndNode="V1" Type="ASL">
  <Content>
   <Field> <!-- This is ASL sign "e" -->
     <Feature>sign</Feature>
     <Value>e</Value>
   </Field></Content>
</AG_Arc>
<AG_Arc ID="A2" StartNode="V3"
  EndNode="V4" Type="Part-of-Speech">
  <Content>VBD</Content>
</AG_Arc>
<AG_Arc ID="A3" StartNode="V0"
  EndNode="V2" Type="Transcription">
  <Content>
   <Field>
     <Feature>AG_Arc</Feature>
     <Value><AG_xref AG_Arc="A2"/></Value>
  </Content>
</AG_Arc>
</AnnotationGraph>
\end{sv}
\end{boxedminipage}
  \caption{AIF: An XML Interchange Format for ATLAS Annotation Graphs}
  \label{fig:aif}
\end{figure}

An example of an interchange format for
structured lexicons is shown in Figure \ref{fig:lexicon}, in this
case using a German to English dictionary as an example.  This example
presents only the most generic type of structure for defining a
lexicon ``signal,'' consisting of an entry for each unique meaning of a
lexical item, and, as in the annotation graph example before, using a
general purpose set of Feature/Value pairs to associate various
features with an item.  As noted in section \ref{sec:gen} on
generalizing the model, ATLAS will enable more special-purpose XML
elements and structure to be derived and used for particular
communities of interest, while retaining the mapping from these back
to a core representation so that all relevant ATLAS-compliant tools
can operate on the data.

\begin{figure}
\begin{boxedminipage}[t]{\linewidth}
\begin{sv}
<AtlasSignal>
 <Signal SignalID="LEX" Class="AtlasLexicon"
    Format="AtlasLexicon:XML" Encoding="XML"
    Comment="German-to-English Dictionary">
  <Entry ID="E1034">
   <Lexeme>reichen</Lexeme>
   <Content>
    <Field>
     <Feature>PartOfSpeech</Feature>
     <!-- Transitive Verb -->
     <Value>VA</Value></Field>
    <Field>
     <Feature>Synonym</Feature>
     <Value>give</Value></Field>
    <Field>
     <Feature>Synonym</Feature>
     <Value>present</Value></Field>
    ...
    <Field>
     <Feature>Idiom</Feature>
     <Value>
      <Field>
       <Feature>Source</Feature>
       <Value>einem die Hand reichen</Value></Field>
      <Field>
       <Feature>Target</Feature>
       <Value>hold out one's hand to someone</Value>
      </Field>
       ...
     </Value></Field></Content>
  </Entry>
  <Entry ID="E1035">
   <Lexeme>reichen</Lexeme>
   <Content>
    <Field>
     <Feature>PartOfSpeech</Feature>
     <!-- Intransitive verb -->
     <Value>VN</Value></Field>
    <Field>
     <Feature>Synonym</Feature>
     <Value>extend to</Value></Field>
    <Field>
     <Feature>Synonym</Feature>
     <Value>suffice</Value></Field>
     ...
   </Entry></Content>
 </Signal>
</AtlasSignal>
\end{sv}
\end{boxedminipage}
  \caption{An XML Interchange Format for Lexicons}
  \label{fig:lexicon}
\end{figure}

\subsection{Implementation -- current status}

At LDC, Michelle Minnick Fox
developed the first implementation of the annotation graph API,
in Perl/tk.  Subsequently, Steven Bird developed a C++
implementation (for both the graph and set APIs).
A team consisting of LDC researchers and programmers
Steven Bird, David Graff, Xiaoyi Ma, Kevin Walker,
Jonathan Wright and Zhibiao Wu
has developed Perl and Tcl interfaces, parsers for a wide
variety of existing corpus formats, and input/output to AIF.
At NIST, Christophe Laprun has developed a Java
implementation of the annotation graph API.

\subsection{Ongoing Activities}

Work is ongoing in several areas at the three sites.
NIST is working with MITRE to implement an ATLAS architecture for the
Automatic Content Extraction (ACE) program Entity Detection and
Tracking (EDT) evaluation, in which Person, Organization,
Geographical-Political, Location, and Facility entities are to be
detected, classified, and clustered \cite{ACE}. The ACE-EDT program will
evaluate the application of EDT technology to text source, audio
source, and image source data.  This multi-domain program will be an
important initial testbed for ATLAS.

NIST is using the ATLAS Interchange Format to implement a new portion
of the NIST Text REtrieval Conference (TREC) Spoken Document Retrieval
Track in which automatic speech recognition systems will be permitted
to output non-lexical information such speaker changes, noise changes,
pauses, etc. in addition to word transcriptions.  These
speech-recognizer-produced files will be exchanged among the research
sites to examine the utility of multiple recognizers and non-lexical
information in performing information retrieval from speech
sources \cite{TREC-9}.

The ATLAS architecture will also be used to provide an information
{\it aqueduct} between components within the DARPA Translingual
Information Detection Extraction and Summarization
program \cite{TIDES}. The TIDES program will examine issues in integrating
multiple language technologies in creating complex multi-media,
multi-lingual information systems.

MITRE has developed two tools that are being ported to use the ATLAS
annotation infrastructure: {\sc Alembic Workbench}, a text annotation
tool that provides mechanisms for annotating lexemes, multi-word
phrases, co-reference relationships and discourse-level structured
entities; and the {\sc Multi-Modal Logger}, which enables multiple
distinct signals (speech, text, application widget events, etc.) to be
aligned and displayed along a common dimension (usually time) and
annotated.  MITRE is also developing a tool to support the declaration
of new ``signal'' classes in ATLAS, enabling customized XML DTDs to be
derived in a principled way such that ATLAS-compliant applications
could operate on these data.

At LDC, development of new annotation tools has been switched over
to follow the ATLAS model, and we envisage a day in the
not-too-distant future when corpora will be released with a
variety of cross-platform tools for accessing and enriching
the corpus content.  We are exploring links with developers of
other tools, including Transcriber \cite{Barras00},
Emu \cite{CassidyHarrington96}, and GATE \cite{Cunningham00},
which have data models that are conceptually close to the
annotation graph model.  A diverse range of
new domains for annotated corpora is being addressed in the
context of the CMU/Penn TalkBank project \cite{TalkBank},
including annotated recordings of animal calls,
and annotated video of sign language.  Relational database
schemas for annotation data are being explored, and a
special purpose query
language is being developed \cite{CassidyBird00,BirdBuneman00}.

All of the code developed within the ATLAS initiative will be
distributed under open source licensing agreements.

\section{Future}

The envisioned ATLAS architecture will support a wide variety of uses
and data types. First and foremost, ATLAS will continue to fulfill its
primary goal of enabling the prodigious exchange of language
resources.  However, in addition to providing support for the internal
and external data representations, it will include integrated access
to useful external resources such as database management systems,
network and security systems, and parsers (Figure~\ref{fig:architecture}).

As the ATLAS architecture is realized, it will enable the rapid
development of new corpora and the exchange and extension of existing
corpora.  It will facilitate the definition of consistent logical and
physical formats for meta data.  It will enable previously-impractical
annotation endeavors requiring multi-domain, multi-layered,
multi-linked annotation.  It will become a focal point for the
development and sharing of modular reusable annotation components and
tools, thus permitting faster application prototyping and development.
These components will provide powerful visualization, manipulation,
and search capabilities over a variety of corpora.

ATLAS will provide a built-in data infrastructure for evaluation by
minimizing the custom-tooling necessary in instrumenting systems for
performance measurement.  The ATLAS Interchange Format will facilitate
the creation of pipelined applications with multiple language
technology components. If desired, however, language technology
applications and components will be able to eliminate pipelining
altogether by using the ATLAS internal representation as a annotation
data exchange bus.

Once the initial instantiation of ATLAS is stabilized, it will be made
into an open-source entity that will benefit from the contribution of
tools and corpora from the worldwide language research community.
While the initial focus is on the annotation of text and speech
corpora, we can envision much broader applications of the architecture
to images, video, non-textual annotation, etc.  Given the open nature
of the project, it is highly likely that it will evolve in ways that
we can't currently predict.  In order to gauge the utility of our
approach and to acquire input on possible uses, the ATLAS working
group has created a public website at:
\sburl{http://www.nist.gov/speech/atlas}.

Before the first public release of ATLAS is made in the Fall of 2000,
the ATLAS working group will be seeking comment from potential users
and developers.  We invite you to peruse the website and send us your
input and ideas.

\bibliographystyle{lrec2000}
\raggedright
\bibliography{general} 

\end{document}